\input ipamacs

\documentstyle[latex-acl]{article}

\title{\vspace{-0.5in}\LARGE\bf SEGMENTING SPEECH WITHOUT A LEXICON:\\
THE ROLES OF PHONOTACTICS AND SPEECH SOURCE}
\author{Timothy Andrew Cartwright \and Michael R. Brent\\
Department of Cognitive Science\\
The Johns Hopkins University\\
3400 North Charles Street\\
Baltimore, MD 21218, USA\\
Internet: {\tt cat@mail.cog.jhu.edu}}

\begin{document}

\maketitle
\vspace{-0.5in}
\begin{abstract}

Infants face the difficult problem of segmenting continuous speech
into words without the benefit of a fully developed lexicon. Several
sources of information in speech might help infants solve this
problem, including prosody, semantic correlations and phonotactics.
Research to date has focused on determining to which of these sources
infants might be sensitive, but little work has been done to determine
the potential usefulness of each source. The computer simulations
reported here are a first attempt to measure the usefulness of
distributional and phonotactic information in segmenting phoneme
sequences. The algorithms hypothesize different segmentations of the
input into words and select the best hypothesis according to the
Minimum Description Length principle. Our results indicate that while
there is some useful information in both phoneme distributions and
phonotactic rules, the combination of both sources is most useful.

\end{abstract}

\section*{INTRODUCTION}

Infants must learn to recognize certain sound sequences as being
words; this is a difficult problem because normal speech contains no
obvious acoustic divisions between words. Two sources of information
that might aid speech segmentation are: distribution---the phoneme
sequence in {\it cat} appears frequently in several contexts including
{\it thecat}, {\it cats} and {\it catnap}, whereas the sequence in
{\it catn} is rare and appears in restricted contexts; and
phonotactics---{\it cat} is an acceptable syllable in English, whereas
{\it pcat} is not. While evidence exists that infants are sensitive to
these information sources, we know of no measurements of their
usefulness. In this paper, we attempt to quantify the usefulness of
distribution and phonotactics in segmenting speech. We found that each
source provided some useful information for speech segmentation, but
the combination of sources provided substantial information. We also
found that child-directed speech was much easier to segment than
adult-directed speech when using both sources.

To date, psychologists have focused on two aspects of the speech
segmentation problem. The first is the problem of parsing continuous
speech into words given a developed lexicon to which incoming sounds
can be matched; both psychologists (e.g., Cutler \& Carter, 1987;
Cutler \& Butterfield, 1992) and designers of speech-recognition
systems (e.g., Church, 1987) have examined this problem. However, the
problem we examined is different---we want to know how infants segment
speech before knowing which phonemic sequences form words. The second
aspect psychologists have focused on is the problem of determining the
information sources to which infants are sensitive. Primarily, two
sources have been examined: prosody and word stress. Results suggest
that parents exaggerate prosody in child-directed speech to highlight
important words (Fernald \& Mazzie, 1991; Aslin, Woodward, LaMendola
\& Bever, in press) and that infants are sensitive to prosody (e.g.,
Hirsh-Pasek et al., 1987). Word stress in English fairly accurately
predicts the location of word beginnings (Cutler \& Norris, 1988;
Cutler \& Butterfield, 1992); Jusczyk, Cutler and Redanz (1993)
demonstrated that 9-month-olds (but not 6-month-olds) are sensitive to
the common strong/weak word stress pattern in English. Sensitivity to
native-language phonotactics in 9-month-olds was recently reported by
Jusczyk, Friederici, Wessels, Svenkerud and Jusczyk (1993). These
studies demonstrated infants' perceptive abilities without
demonstrating the usefulness of infants' perceptions.

How do children combine the information they perceive from different sources?
Aslin et al.\
speculate that infants first learn words heard in isolation, then use
distribution and prosody
to refine and expand their vocabulary; however, Jusczyk (1993) suggests that
sound sequences
learned in isolation differ too greatly from those in context to be useful. He
goes on to say,
``just how far information in the sound structure of the input can bootstrap
the acquisition
of other levels [of linguistic organization] remains to be determined.'' In
this paper, we
measure the potential roles of distribution, phonotactics and their combination
using a
computer-simulated learning algorithm; the simulation is based on a
bootstrapping model in
which phonotactic knowledge is used to constrain the distributional analysis of
speech
samples.

While our work is in part motivated by the above research, other developmental
research
supports certain assumptions we make. The input to our system is represented as
a sequence of
phonemes, so we implicitly assume that infants are able to convert from
acoustic input to
phoneme sequences; research by Kuhl (e.g., Grieser \& Kuhl, 1989) suggests that
this
assumption is reasonable. Since sentence boundaries provide information about
word boundaries
(the end of a sentence is also the end of a word), our input contains sentence
boundaries;
several studies (Bernstein-Ratner, 1985; Hirsh-Pasek et al., 1987; Kemler
Nelson, Hirsh-Pasek,
Jusczyk \& Wright Cassidy, 1989; Jusczyk et al., 1992) have shown that infants
can perceive
sentence boundaries using prosodic cues. However, Fisher and Tokura (in press)
found no
evidence that prosody can accurately predict word boundaries, so the task of
finding words
remains. Finally, one might question whether infants have the ability we are
trying to
model---that is, whether they can identify words embedded in sentences; Jusczyk
and Aslin
(submitted) found that 7 1/2-month-olds can do so.

\subsection*{The Model}

To gain an intuitive understanding of our model, consider the following speech
sample
(transcription is in IPA):
\begin{center}
\begin{tabular}{rl}
Orthography: & Do you see the kitty?\\
             & See the kitty?\\
             & Do you like the kitty?\\[1.5 ex]
Transcription: & dujusi\eth \schwa k\sci ti\\
               & si\eth \schwa k\sci ti\\
               & dujula\sci k\eth \schwa k\sci ti
\end{tabular}
\end{center}
There are many different ways to break this sample into putative words (each
particular
segmentation is called a segmentation hypothesis). Two such hypotheses are:
\begin{center}
\begin{tabular}{rl}
Segmentation 1: & du ju si \eth \schwa\ k\sci ti\\
                & si \eth \schwa\ k\sci ti\\
                & du ju la\sci k \eth \schwa\ k\sci ti\\[1.5 ex]
Segmentation 2: & duj us i\eth\ \schwa k\sci t i\\
                & si\eth\ \schwa k \sci ti\\
                & du jul a\sci k \eth \schwa k \sci ti
\end{tabular}
\end{center}
Listing the words used by each segmentation hypothesis yields the following two
lexicons:
\begin{center}
\begin{tabular}{r@{  }lr@{  }lr@{  }l}
\multicolumn{6}{c}{Segmentation 1} \\ \hline
1 & du         & 3 & k\sci ti & 5 & si \\
2 & \eth\schwa & 4 & la\sci k & 6 & ju \\[2ex]
\multicolumn{6}{c}{Segmentation 2} \\ \hline
1 & a\sci k      & 5 & \schwa k       &  9 & \sci ti \\
2 & du           & 6 & \schwa k\sci t & 10 & jul     \\
3 & duj          & 7 & i              & 11 & si\eth  \\
4 & \eth\schwa k & 8 & i\eth          & 12 & us
\end{tabular}
\end{center}
Note that Segmentation 1, the correct hypothesis, yields a compact lexicon of
frequent words
whereas Segmentation 2 yields a much larger lexicon of infrequent words. Also
note that a
lexicon contains only the words used in the sample---no words are known to the
system a
priori, nor are any carried over from one hypothesis to the next. Given a
lexicon, the sample
can be encoded by replacing words with their respective indices into the
lexicon:
\begin{center}
\begin{tabular}{rl}
Encoded Sample 1: & 1, 6, 5, 2, 3; \\
                  & 5, 2, 3; \\
                  & 1, 6, 4, 2, 3; \\[1.5ex]
Encoded Sample 2: & 2, 12, 6, 4, 5; \\
                  & 11, 3, 8; \\
                  & 1, 9, 10, 7, 8;
\end{tabular}
\end{center}
Our simulation attempts to find the hypothesis that minimizes the combined
sizes of the
lexicon and encoded sample. This approach is called the Minimum Description
Length (MDL)
paradigm and has been used recently in other domains to analyze distributional
information (Li
\& Vit\'{a}nyi, 1993; Rissanen, 1978; Ellison, 1992, 1994; Brent, 1993). For
reasons explained
in the next section, the system converts these character-based representations
to compact
binary representations, using the number of bits in the binary string as a
measure of size.

Phonotactic rules can be used to restrict the segmentation hypothesis space by
preventing word
boundaries at certain places; for instance, /k{\ae}tsp\openo z/ (``cat's
paws'') has six
internal segmentation points (k {\ae}tsp\openo z, k{\ae} tsp\openo z, etc),
only two of which
are phonotactically allowed (k{\ae}t~sp\openo z and k{\ae}ts~p\openo z). To
evaluate the
usefulness of phonotactic knowledge, we compared results between
phonotactically constrained
and unconstrained simulations.

\section*{SIMULATION DETAILS}

To use the MDL principle, as introduced above, we search for the smallest-sized
hypothesis. We
must have some well-defined method of measuring hypothesis sizes for this
method to work. A
simple, intuitive way of measuing the size of a hypothesis is to count the
number of
characters used to represent it. For example, counting the characters
(excluding spaces) in
the introductory examples, we see that Hypothesis 1 uses 48 characters and
Hypothesis 2 uses
75. However, this simplistic method is inefficient; for instance, the length of
lexical
indices are arbitrary with respect to properties of the words themselves (e.g.,
in Hypothesis
2, there is no reason why /jul/ was assigned the index `10'---length
two---instead of
`9'---length one). Our system improves upon this simple size metric by
computing sizes based
on a compact representation motivated by information theory.

We imagine hypotheses represented as a string of ones and zeros. This binary
string must
represent not only the lexical entries, their indices (called {\bf code words})
and the coded
sample, but also overhead information specifying the number of items coded and
their
arrangement in the string (information implicitly given by spacing and spatial
placement in
the introductory examples). Furthermore, the string and its components must be
self-delimiting, so that a decoder could identify the endpoints of components
by itself. The
next section describes the binary representation and the length formul{\ae}
derived from it in
detail; readers satisfied with the intuitive descriptions presented so far
should skip ahead
to the Phonotactics sub-section.

\subsection*{Representation and Length Formul{\ae}}

The representation scheme described below is based on information theory (for
more examples of
coding systems, see, e.g., Li \& Vit\'{a}nyi, 1993 and Quinlan \& Rivest,
1989). From this
representation, we can derive a formula describing its length in bits. However,
the discrete
form of the formula would not work well in practice for our simulations.
Instead, we use a
continuous approximation of the discrete formula; this approximation typically
involves
dropping the ceiling function from length computations. For example, we
sometimes use a
self-delimiting representation for integers (as described in Li \& Vit\'{a}nyi,
pp. 74--75). In this representation, the number of bits needed to code an
integer $x$ is given
by
\[
\ell^{(2)}(x) = 1 + \left\lceil\log_2(x+1)\right\rceil +
2\left\lceil\log_2\left\lceil\log_2(x+1)\right\rceil\right\rceil
\]
However, we use the following approximation:
\[
\ell^{(2)}(x) = 1.5 + \log_2(x+1) + 2\log_2(\log_2(x+2)+0.5)
\]
Using the discrete formula, the difference between $\ell^{(2)}(126)$ and
$\ell^{(2)}(127)$ is
zero, while the difference between $\ell^{(2)}(127)$ and $\ell^{(2)}(128)$ is
one bit; using
the continuous formula, the difference between $\ell^{(2)}(126)$ and
$\ell^{(2)}(127)$ is
0.0156, while the difference between $\ell^{(2)}(127)$ and $\ell^{(2)}(128)$ is
0.0155. We
found it easier to interpret the results using a continuous function, so in the
following
discussion, we will only present the approximate formul{\ae}.

The lexicon lists words (represented as phoneme sequences) paired with their
code
words\footnote{Code words are represented by square brackets, so [$x$] means
`the code word
corresponding to $x$'.}. For example:
\begin{center}
\begin{tabular}{cc}
Word & Code Word \\ \hline
\eth\schwa & [the] \\
k{\ae}t    & [cat] \\
k\sci ti   & [kitty] \\
si         & [see] \\
$\vdots$   & $\vdots$
\end{tabular}
\end{center}
In the binary representation, the two columns are represented separately, one
after the other;
the first column is called the {\bf word inventory column}; the second column
is called the
{\bf code word inventory column}.

\begin{figure*}
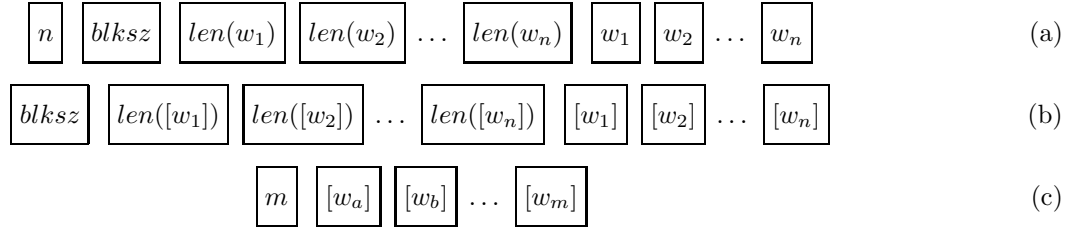

\begin{center}
\caption{\label{box_diagram}\it
Schematic diagrams for components of the representation}
\vspace{0.1in}
\hfill
\fbox{\rule[-5pt]{0cm}{16pt}$n$}\hspace{4pt}
\fbox{\rule[-5pt]{0cm}{16pt}$blksz$}\hspace{4pt}
\fbox{\rule[-5pt]{0cm}{16pt}$len(w_{1})$}\hspace{2pt}
\fbox{\rule[-5pt]{0cm}{16pt}$len(w_{2})$}\hspace{1pt}
\ldots\hspace{1pt}
\fbox{\rule[-5pt]{0cm}{16pt}$len(w_{n})$}\hspace{4pt}
\fbox{\rule[-5pt]{0cm}{16pt}$w_{1}$}\hspace{2pt}
\fbox{\rule[-5pt]{0cm}{16pt}$w_{2}$}\hspace{1pt}
\ldots\hspace{1pt}
\fbox{\rule[-5pt]{0cm}{16pt}$w_{n}$}
\hfill
(a)\\[0.1in]
\hfill
\fbox{\rule[-5pt]{0cm}{16pt}$blksz$}\hspace{4pt}
\fbox{\rule[-5pt]{0cm}{16pt}$len([w_{1}])$}\hspace{2pt}
\fbox{\rule[-5pt]{0cm}{16pt}$len([w_{2}])$}\hspace{1pt}
\ldots\hspace{1pt}
\fbox{\rule[-5pt]{0cm}{16pt}$len([w_{n}])$}\hspace{4pt}
\fbox{\rule[-5pt]{0cm}{16pt}$[w_{1}]$}\hspace{2pt}
\fbox{\rule[-5pt]{0cm}{16pt}$[w_{2}]$}\hspace{1pt}
\ldots\hspace{1pt}
\fbox{\rule[-5pt]{0cm}{16pt}$[w_{n}]$}
\hfill
(b)\\[0.1in]
\hfill
\fbox{\rule[-5pt]{0cm}{16pt}$m$}\hspace{4pt}
\fbox{\rule[-5pt]{0cm}{16pt}$[w_{a}]$}\hspace{2pt}
\fbox{\rule[-5pt]{0cm}{16pt}$[w_{b}]$}\hspace{1pt}
\ldots\hspace{1pt}
\fbox{\rule[-5pt]{0cm}{16pt}$[w_{m}]$}
\hfill
(c)
\end{center}
\end{figure*}

In the word inventory column (see Figure \ref{box_diagram}a for a schematic),
the list of
lexical items is represented as a continuous string of phonemes, without
separators between
words (e.g., \eth\schwa k{\ae}tk\sci tisi\ldots). To mark the boundaries
between lexical
items, the phoneme string is preceded by a list of integers representing the
lengths (in
phonemes) of each word. Each length is represented as a fixed-length,
zero-padded binary
number. Preceding this list is a single integer denoting the length of each
length field; this
integer is represented in unary, so that its length need not be known in
advance. Preceding
the entire column is the number of lexical entries $n$ coded as a
self-delimiting integer.

The length of the representation of the integer $n$ is given by the function
\begin{equation}
\ell^{(2)}(n)
\label{wi_n}
\end{equation}

We define $len(w_{i})$ to be the number of phonemes in word $w_i$. If there are
$p$ total
unique phonemes used in the sample, then we represent each phoneme as a
fixed-length bit
string of length $len(p) = \log_2 p$. So, the length of the representation of a
word $w_{i}$
in the lexicon is the number of phonemes in the word times the length of a
phoneme: $len(p)
\cdot len(w_{i})$. The total length of all the words in the lexicon is the sum
of this formula
over all lexical items:
\begin{equation}
\sum_{i=1}^n \left( len(p) \cdot len(w_i) \right) = len(p) \sum_{i=1}^n
len(w_i)
\label{wi_words}
\end{equation}

As stated above, the length fields used to divide the phoneme string are
fixed-length. In each
field is an integer between one and the number of phonemes in the longest word.
Since
representing integers between one and $x$ takes $\log_2 x$ bits, the length of
each field is:
\[
\log_2(\max_{1...n}\;len(w_i))
\]
To be fully self-delimiting, the width of a field must be represented in a
self-delimiting
way; we use a unary representation---i.e., write an extra field consisting of
only `1' bits
followed by a terminating `0'. There are $n$ fields (one for each word), plus
the unary
prefix, so the combined length of the fields plus prefix (plus terminating
zero) is:
\begin{equation}
1 + (n+1)\,\log_2(\max_{1...n}\;len(w_i))
\label{wi_prefix}
\end{equation}
The total length of the word inventory column representation is the sum of the
terms in
(\ref{wi_n}), (\ref{wi_words}) and (\ref{wi_prefix}).

The code word inventory column of the lexicon (see Figure \ref{box_diagram}b
for a schematic)
has a nearly identical representation as the previous column except that code
words are listed
instead of phonemic words---the length fields and unary prefix serve the same
purpose of
marking the divisions between code words.

The sample can be represented most compactly by assigning short code words to
frequent words,
reserving longer code words for infrequent words. To satisfy this property,
code words are
assigned so that their lengths are frequency-based; the length of the code word
for a word of
frequency $f(w)$ will not be greater than:
\[
len([w]) = \log_2 \frac{\sum_{i=1}^n f(w)}{f(w)} = \log_2 \frac{m}{f(w)}
\]
The total length of the code word list is the sum of the code word lengths over
all lexical
entries:
\begin{equation}
\sum_{i=1}^n len([w]) = \sum_{i=1}^n \log_2\frac{m}{f(w_i)}
\label{cwi_codes}
\end{equation}
As in the word inventory column (described above), the length of each code word
is represented
in a fixed-length field. Since the least frequent word will have the longest
code word (a
property of the formula for $len([w_i])$), the longest possible code word comes
from a word of
frequency one:
\[
\log_2\frac{m}{1} = \log_2 m
\]
Since the fields contains integers between one and this number, we define the
length of a
field to be:
\[
\log_2 ( \log_2 m )
\]
As above, we represent the width of a field in unary, so there are a total of
$n+1$ elements
of this size ($n$ fields plus the unary representation of the field width). The
combined
length of the fields plus prefix (and terminating zero) is:
\begin{equation}
1 + (n+1) \log_2 ( \log_2 m )
\label{cwi_prefix}
\end{equation}
The total length of the code word inventory column representation is the sum of
the terms in
(\ref{cwi_codes}) and (\ref{cwi_prefix}).

Finally, the sequence of words which form the sample (see Figure
\ref{box_diagram}c for a
schematic) is represented as the number of words in the sample ($m$) followed
by the list of
code words. Since code words are used as compact indices into the lexicon, the
original sample
could be reconstructed completely by looking up each code word in this list and
replacing it
with its phoneme sequence from the lexicon. The code words we assigned to
lexical items are
self-delimiting (once the set of codes is known), so there is no need to
represent the
boundaries between code words.

The length of the representation of the integer $m$ is given by the function
\begin{equation}
\ell^{(2)}(m)
\label{cs_m}
\end{equation}

The length of the representation of the sample is computed by summing the
lengths of the code
words used to represent the sample. We can simplify this description by noting
that the
combined length of all occurrences of a particular code word $[w_i]$ is
$f(w_i)\cdot
len([w_i])$ since there are $f(w_i)$ occurrences of the code word in the
sample. So, the
length of the encoded sample is the sum of this formula over all words in the
{\bf lexicon}:
\begin{equation}
\sum_{i=1}^n f(w_i) \cdot len([w]) =
\sum_{i=1}^n \left[ f(w_i) \cdot \log_2 \left( \frac{m}{f(w_i)} \right) \right]
\label{cs_words}
\end{equation}
The total length of the sample is given by adding the terms in (\ref{cs_m}) and
(\ref{cs_words}). The total length of the representation of the entire
hypothesis is the sum
of the representation lengths of the word inventory column, the code word
inventory column and
the sample.

This system of computing hypothesis sizes is efficient in the sense that
elements are thought
of as being represented compactly and that code words are assigned based on the
relative
frequencies of words. The final evaluation given to a hypothesis is an estimate
of the minimal
number of bits required to transmit that hypothesis. As such, it permits direct
comparison
between competing hypotheses; that is, the shorter the representation of some
hypothesis, the
more distributional information can be extracted and, therefore, the better the
hypothesis.

\subsection*{Phonotactics}

Phonotactic knowledge was given to the system as a list of licit initial and
final consonant
clusters of English words\footnote{In phonological terms, the syllable onsets
permitted at
word beginnings and syllable codas permitted at word ends. Some languages
(including English)
have different sets of onsets and codas for word-internal and word-boundary
positions---we use
the word-boundary set.}; this list was checked against all six samples so that
the list was
maximally permissive (e.g., the underlined consonant cluster in
e\underline{xpl}ore could be
divided as ek-splore or eks-plore). In those simulations which used the
phonotactic knowledge,
a word boundary could not be inserted when doing so would create a word initial
or final
consonant cluster not on the list or would create a word without a vowel. For
example (from an
actual sample---corresponds to the utterance, ``Want me to help baby?''):
\begin{center}
\begin{tabular}{rl}
Sample: & wantmituh\niepsilon lpbebi \\
Valid Boundaries: & want.mi.t.u.h\niepsilon lp.be.bi
\end{tabular}
\end{center}
In the second line, those word boundaries that are phonotactically legal are
marked with
dots. The boundary between /w/ and /a/ is illegal because /w/ by itself is not
a legal word in
English; the boundary between /a/ and /n/ is illegal because /ntm/ is not a
valid word initial
consonant cluster; the boundary between /m/ and /i/ is illegal because /ntm/ is
also not a
valid word final consonant cluster; the boundary between /p/ and /b/ is legal
because /lp/ is
a valid word final cluster and /b/ is a valid word initial cluster. Note that
using the
phonotactic constraints reduces the number of potential word boundaries from
fifteen to six in
this example.

After the system inserts a new word boundary, it updates the list of remaining
valid insertion
points---adding a point may cause nearby points to become unusable due to the
restriction that
every word must have a vowel. For example (corresponding to the utterance
``green and''):
\begin{center}
\begin{tabular}{rl}
Before: & gri.n.{\ae}nd \\
After: & grin \\
& {\ae}nd
\end{tabular}
\end{center}
After the segmentation of /grin/ and /{\ae}nd/, the potential boundary between
/i/ and /n/
becomes invalid because inserting a word boundary there would produce a word
with no vowel
(/n/).

\subsection*{Inputs and Simulations}

Two speech samples from each of three subjects were used in the simulations; in
one sample a
mother was speaking to her daughter and in the other, the same mother was
speaking to the
researcher. The samples were taken from the CHILDES database (MacWhinney \&
Snow, 1990) from
studies reported in Bernstein (1982). Each sample was checked for consistent
word spellings
(e.g., 'ts was changed to its), then was transcribed into an ASCII-based
phonemic
representation\footnote{The transcription method ensured the identical
transcription of all
occurrences of a word.}. The transcription system was based on IPA and used one
character for
each consonant or vowel; diphthongs, r-colored vowels and syllabic consonants
were each
represented as one character. For example, ``boy'' was written as {\tt b7},
``bird'' as {\tt
bRd} and ``label'' as {\tt lebL}. For purposes of phonotactic constraints,
syllabic consonants
were treated as vowels. Sample lengths were selected to make the number of
available
segmentation points nearly equal (about 1,350) when no phonotactic constraints
were applied;
child-directed samples had 498--536 tokens and 153--166 types, adult-directed
samples had
443--484 tokens and 196--205 types. Finally, before the samples were fed to the
simulations,
divisions between words (but not between sentences) were removed.

The space of possible hypotheses is vast\footnote{For our samples,
unconstrained by
phonotactics, there are about $2^{1350} \approx 2.5 \times 10^{406}$
hypotheses.}, so some
method of finding a minimum-length hypothesis without considering all
hypotheses is
necessary. We used the following method: first, evaluate the input sample with
no segmentation
points added; then evaluate all hypotheses obtained by adding one or two
segmentation points;
take the shortest hypothesis found in the previous step and evaluate all
hypotheses obtained
by adding one or two more segmentation points; continue this way until the
sample has been
segmented into the smallest possible units and report the shortest hypothesis
ever found.  Two
variants of this simulation were used: (1) {\sc Dist-Free} was free of any
phonotactic
restrictions on the hypotheses it could form ({\sc Dist} refers to the
measurement of
distributional information), whereas (2) {\sc Dist-Phono} used the phonotactic
restrictions
described above. Each simulation was run on each sample, for a total of twelve
{\sc Dist}
runs.

Finally, two other simulations were run on each sample to measure chance
performance: (1) {\sc
Rand-Free} inserted random segmentation points and reported the resulting
hypothesis, (2) {\sc
Rand-Phono} inserted random segmentation points where permitted by the
phonotactic
constraints. Since the {\sc Rand} simulations were given the number of
segmentation points to
add (equal to the number of segmentation points needed to produce the natural
English
segmentation), their performance is an upper bound on chance performance. In
contrast, the
{\sc Dist} simulations must determine the number of segmentation points to add
using MDL
evaluations. The results for each {\sc Rand} simulation are averages over 1,000
trials on each
input sample.

\section*{RESULTS}

Each simulation was scored for the number of correct segmentation points
inserted, as compared
to the natural English segmentation. From this scoring, two values were
computed: {\bf
recall}, the percent of all correct segmentation points that were actually
found; and {\bf
accuracy}, the percent of the hypothesized segmentation points that were
actually correct. In
terms of hits, false alarms and misses, we have:
\begin{eqnarray*}
recall   & = & \frac{hits}{hits+misses}\\
accuracy & = & \frac{hits}{hits+false\ alarms}
\end{eqnarray*}

\begin{table*}
\begin{center}
\caption{\label{res_table}\it
Results for all simulations averaged over individual speech samples\\}
\renewcommand{\arraystretch}{1.5}
\begin{tabular}{llcccc} \hline
 & & \multicolumn{4}{c}{Simulation} \\ \cline{3-6}
Target & Measure & \sc Rand-Free & \sc Rand-Phono & \sc Dist-Free & \sc
Dist-Phono \\ \hline
Adult   & \% Recall   & 25.1 & 39.5 & 95.5 & 22.5 \\
        & \% Accuracy & 28.9 & 50.5 & 36.0 & 92.0 \\[1ex]
Child   & \% Recall   & 23.4 & 40.2 & 79.9 & 72.3 \\
        & \% Accuracy & 26.7 & 51.7 & 37.4 & 88.3 \\[1ex] \hline
Average & \% Recall   & 24.3 & 39.9 & 88.0 & 46.4 \\
        & \% Accuracy & 27.8 & 51.1 & 36.6 & 89.2 \\ \hline
\end{tabular}
\end{center}
\end{table*}

Results are given in Table \ref{res_table}. Note that there is a trade-off
between recall and
accuracy---if all possible segmentation points were added, recall would be
100\% but accuracy
would be low; likewise, if only one segmentation point was added between two
words, accuracy
would be 100\% but recall would be low. Since our goal is to correctly segment
speech,
accuracy is more important than finding every correct segmentation. For
example, deciding
`littlekitty' is a word is less disastrous than deciding `li', `tle', `ki' and
`ty' are all
words, because assigning meaning to `littlekitty' is a reasonable first try at
learning
word-meaning pairs, whereas trying to assign separate meanings to `li' and
`tle' is
problematic.

The performance of {\sc Dist-Phono} on child-directed speech shows that this
system goes a
long way toward solving the segmentation problem. However, comparing the
average performances
of simulations is also useful. The effect of phonotactic information can be
seen by comparing
the average performances of {\sc Rand-Free} and {\sc Rand-Phono}, since the
only difference
between them is the addition of phonotactic constraints on segmentations in the
latter. Clearly phonotactic constraints are useful, as both recall and accuracy
improve. A
similar comparison between {\sc Rand-Free} and {\sc Dist-Free} shows that
distributional
information alone also improves performance.  Note in all the results of {\sc
Dist-Free} that
using distributional information alone favors recall over accuracy; in fact,
the segmentation
hypotheses produced by {\sc Dist-Free} have most words broken into single
phoneme units with
only a handful of words remaining intact. Two comparisons are needed to show
that the
combination of distributional and phonotactic information performs better than
either source
alone: {\sc Dist-Phono} compared to {\sc Rand-Phono}, to see the effect of
adding
distributional analysis to phonotactic constraints, and {\sc Dist-Phono}
compared to {\sc
Dist-Free}, to see the effect of adding phonotactic constraints to
distributional
analysis. The former comparison shows that the sources combined are more useful
than
phonotactic information alone. The latter comparison is less obvious---the
trade-off between
recall and accuracy seems to have reversed, with no clear winner\footnote{The
higher accuracy
of {\sc Dist-Phono} is a good sign. Furthermore, the minimum of the
recall/accu- racy pair is
greater in {\sc Dist-Phono} than in {\sc Dist-Free} and the maximum of the
recall/accuracy
pair is also greater in {\sc Dist-Phono} than in {\sc Dist-Free}.}. Data on
discovered word
types helps make this comparison: {\sc Dist-Free} found 12\% of the words with
30\% accuracy
and {\sc Dist-Phono} found 33\% of the words with 50\% accuracy. Whereas the
segmentation
point data are inconclusive, word type data demonstrate that combining
information sources is
more useful than using distributional information alone.

There is no obvious difference in performance between child- and adult-directed
speech, except
in {\sc Dist-Phono} (combined information sources) in which the difference is
striking:
accuracy remains high and recall rate more than triples for child-directed
speech. This
difference is again supported by word type data: 14\% recall with 30\% accuracy
for
adult-directed speech, 56\% recall with 65\% accuracy for child-directed
speech.

\section*{DISCUSSION}

Our technique segments continuous speech into words using only distributional
and phonotactic
information more effectively than one might expect---up to 66\% recall of
segmentation points
with 92\% accuracy on one sample, which yields 58\% recall of word types with
67\% accuracy
(the relatively low type accuracy is mitigated by the fact that most incorrect
words are
meaningful concatenations of correct words---e.g., `thekitty'). This finding
confirms the idea
that distribution and phonotactics are useful sources of information that
infants might use in
discovering words (e.g., Jusczyk et al., 1993b). In fact, it helps explain
infants' ability to
learn words from parental speech: these two sources alone are useful and
infants have several
others, like prosody and word stress patterns, available as well. It also
suggests that
semantics and isolated words need not play as central a role as one might think
(e.g.,
Jusczyk, 1993, downplayed the utility of words in isolation). It is difficult,
if not
impossible given currently available methods, to determine which sources of
information are
necessary for infants to segment speech and learn words; only this sort of
indirect evidence
is available to us.

The results show a difference between adult- and child-directed speech, in that
the latter is
easier to segment given both distribution and phonotactics. This lends
quantitative support to
research which suggests that motherese differs from normal adult speech in ways
possibly
useful to the language-learning infant (Aslin et al.). In fact, the factors
making motherese
more learnable might be elucidated using this technique: compare the results of
several
different models, each containing a different factor or combination of factors,
looking for
those in which a substantial performance difference exists between child- and
adult-directed
speech.

Our model uses phonotactic constraints as absolute requirements on the
structure of individual
words; this implies that phonotactics have been learned prior to attempts at
segmentation. We
must therefore show that phonotactics can indeed be learned without access to a
lexicon---without such a demonstration, we are trapped in circular reasoning.
Gafos and Brent
(1994) demonstrate that phonotactics can be learned with high accuracy from the
same
unsegmented utterances we used in our simulations. In general, two methods
exist for combining
information sources in the MDL paradigm: one is to have absolute requirements
on plausible
hypotheses (like our phonotactic constraints)---these requirements must be
independently
learnable; the other method of combination is to include an information source
in the internal
representation of hypotheses (like our distributional information)---all
components of the
representation are learned simultaneously (see Ellison, 1992, for an example of
multiple
components in a representation).

We would like to extend the system by using a more detailed transcription
system. We expect
that this would help the system find word boundaries for reasons detailed in
Church
(1987)---in brief, that allophonic variation may be quite useful in predicting
word
boundaries. Another simpler extension of this research will be to increase the
length of the
speech samples used. Finally, we will try the current system on samples from
other languages,
to make sure this method generalizes appropriately.

This research program will provide complementary evidence supporting hypotheses
about the
sources of information infants use in learning their native languages. Until
now, research has
focused on demonstrations of infants' sensitivity to various sources; we have
begun to provide
quantitative measures of the usefulness of those sources.

\end{document}